\documentclass[submission,copyright,creativecommons]{eptcs}
\providecommand{\event}{ICLP DC 2020} % Name of the event you are submitting to
\usepackage{epsfig}

\usepackage{todonotes}

\title{Constraint Programming Algorithms for Route Planning Exploiting Geometrical Information}
\author{Alessandro Bertagnon
\institute{Department of Engineering\\
University of Ferrara\\
Ferrara, Italy}
\email{alessandro.bertagnon@unife.it}
}
\def\titlerunning{Constraint Programming Algorithms for Route Planning Exploiting Geometrical Information}
\def\authorrunning{A. Bertagnon}%R.J. van Glabbeek, C. Author \& Y.S. Else}

\usepackage{acronym}
\newacro{TSP}{Traveling Salesperson Problem}
\newacro{TSPTW}{TSP with Time Windows}
\newacro{CP}{Constraint Programming}
\newacro{CSP}{Constraint Satisfaction Problem}
\newacro{COP}{Constraint Optimization Problem}
\newacro{PTAS}{Polynomial Time Approximation Scheme}
\newacro{VRP}{Vehicle Routing Problem}
\newacro{CLP}{Constraint Logic Programming}
\newacro{WCC}{Weighted Circuit Constraint}
\newacro{ASP}{Answer Set Programming}

\usepackage{amsmath}
\usepackage{amssymb}
\usepackage{amsthm}
\usepackage{graphicx}

\newcommand{\point}{{\ensuremath{P}}}

\newcommand{\segment}[2]{{\ensuremath{\overline{#1 #2}}}}
\newcommand{\StraightLine}[2]{{\ensuremath{\overleftrightarrow{#1 #2}}}} % straight line passing through two points (retta)
\newcommand{\OptLength}{{\ensuremath{L^*}}}
\newcommand{\Npoints}{{\ensuremath{n}}}	% Number of points

\newcommand{\Next}{{\ensuremath{\mathit{Next}}}}	% Domain variables in the successor representation
\newcommand{\nocrossing}{{\tt nocrossing}}
\newcommand{\clockwise}{{\tt clockwise}}
\newcommand{\Dom}[1]{{\ensuremath{\mathit{Dom}(#1)}}}

\newcommand{\Angle}[1]{{\ensuremath{\angle #1}}}
\newcommand{\Hull}{{\ensuremath{H}}}
\newcommand{\circuit}{{\tt circuit}}
\newcommand{\alldifferent}{{\tt alldifferent}}

\newcommand{\minAlpha}{\ensuremath{\underline{\alpha}}}
\newcommand{\maxBeta}{\ensuremath{\overline{\beta}}}

\newcommand{\ECLiPSe}{ECL$^i$PS$^e$}
\newcommand{\noi}{Geometric}
\newcommand{\loro}{BvHRRR}
\newcommand{\noipiuloro}{Geom.+BvHRRR}
\newcommand{\eclipseBase}{CLP(FD)}

\newtheorem{theorem}{Theorem}
\newtheorem{property}{Property}
\newtheorem{definition}{Definition}

\usepackage[caption=false]{subfig}
\usepackage{pgfplots}
\usetikzlibrary{plotmarks}
\usepgfplotslibrary{groupplots}
\usetikzlibrary{matrix}
\pgfplotsset{compat=newest}
\usepackage{tikz}
\usetikzlibrary{decorations}
\usetikzlibrary{decorations.markings}

\usepackage{floatrow}
\begin{document}
\maketitle

\begin{abstract}
Problems affecting the transport of people or goods are plentiful in industry and commerce and they also appear to be at the origin of much more complex problems. In recent years, the logistics and transport sector keeps growing supported by technological progress, i.e. companies to be competitive are resorting to innovative technologies aimed at efficiency and effectiveness. This is why companies are increasingly using technologies such as Artificial Intelligence (AI), Blockchain and Internet of Things (IoT). 
%Artificial intelligence in particular plays a fundamental role in providing, among other things, solutions to problems capable of optimize the use of the available resources.
Artificial intelligence, in particular, is often used to solve optimization problems in order to provide users with the most efficient ways to exploit available resources.
%\todo{Non molto chiaro. chi \`e il soggetto di "Capable"? Immagino "solutions", ma non \`e fluido.}

In this work we present an overview of our current research activities concerning the development of new algorithms, based on \ac{CLP} techniques, for route planning problems exploiting the geometric information intrinsically present in many of them or in some of their variants.
The research so far has focused in particular on the Euclidean Traveling Salesperson Problem (Euclidean TSP) with the aim to exploit the results obtained also to other problems of the same category, such as the Euclidean Vehicle Routing Problem (Euclidean VRP), in the future.

\end{abstract}

\section{Introduction}
Given a weighted graph $G$ with $n$ vertices the \ac{TSP} requires to compute the shortest cycle that visits each vertex of $G$ exactly once \footnote{A cycle that visit each vertex exactly once is often referred as Hamiltonian cycle or Hamiltonian circuit.}. The name ``Traveling Salesman Problem" comes from the problem's most famous formulation: ``A salesman has to visit a set of cities, each of which must be visited only once, and he wants to minimize the length of the tour". The problem is NP-hard \cite{Karp72}. 
Currently, the best solver for the \ac{TSP} is Concorde \cite{concorde}, that includes several techniques based on Integer Linear Programming (with branch-and-bound and branch-and-cut) and Local Search. 

Some significant sub-classes of the general \ac{TSP} are the {\em metric TSP}, in which the distance function between cities enjoys the triangle inequality, and the {\em Euclidean TSP}, in which the nodes of the graph represent points in the plane and the distance function
is the Euclidean distance. These are reasonable assumptions in many important instances: many industrial problem and various benchmarks taken from the TSPLIB \cite{tsplib} fall into these classes.%\todo{in questa frase "instances" c'\`e 3 volte.}

Both the metric and the Euclidean TSP, as the general \ac{TSP}, are NP-hard \cite{TSPeuclideoNPhard};
nonetheless, differently from the general \ac{TSP}, the Euclidean \ac{TSP} admits a \ac{PTAS} \cite{Arora_PTAS,Mitchell_PTAS},
i.e., given a value $\epsilon >0$, it is possible to obtain  a solution with cost $(1+\epsilon)\OptLength$ (where \OptLength\ is the length of the optimal tour) in polynomial time with respect to the number of nodes \Npoints\ (note, however, that the time is exponential with respect to $\frac{1}{\epsilon}$). 

It is worth noting that in the Euclidean TSP more information is available than in the general \ac{TSP}: the coordinates of the points to be visited are known, and geometrical concepts (straight line segments, angles, etc.) can be defined in the Euclidean plane.
Despite these results are very important theoretically, \ac{TSP} solvers do not use the additional geometric information which is available in the Euclidean TSP instances. Concorde, which is in practice faster in most applications, discards the information regarding vertices position on the plane and it computes the distance matrix using Euclidean distances as weights.

In the literature another important work that attempted to use geometric information has recently been published: Deudon et al. \cite{DBLP:conf/cpaior/DeudonCLAR18} train a Deep Neural Network with points coordinates to learn efficient heuristics to explore the search space. This work, instead, is the first attempt (to the best of our knowledge) to exploit geometric information to obtain further pruning in \ac{CP} during the solution of some route planning problems.

\section{Related Works}
As mentioned in the previous section, the best solver currently available for the \ac{TSP} is Concorde\cite{concorde}; but Concorde can
address only {\em pure} \ac{TSP}s, i.e., no further side constraints are allowed, while in \ac{CP} many variants can be easily cast, such as the \ac{TSPTW}.

Three representations have been devised, in \ac{CP} literature, for defining variables in the Hamiltonian circuit problem and the TSP: the {\em permutation} representation, the {\em successor} representation and the {\em set variable} representation \cite{BenchimolHRRR12} later extended to the graph representation \cite{CP_graph,DBLP:journals/corr/abs-1206-3437,SalesmanAndTree}. In the following of this paper, we will be mainly concerned with the successor representation.

Given a list $L$ of variables $\left[ \Next_{1},\Next_{2},...,\Next_{n} \right]$, where $n$ is the number of vertices of the graph $G$, we denote with $\Dom{\Next_{i}}$
the initial domain of the variable $\Next_{i}$, where $\Dom{\Next_{i}} = \{1,\dots,\Npoints\} \setminus \{i\}$.
In the successor representation, the value of the $\Next_{i}$ variable denotes the successor of the vertex $i$ in the resulting tour (e.g. if $n=5$ and $L = \left[ 3,5,4,2,1\right]$ the corresponding tour will be $1,3,4,2,5,1$). Note that in the Euclidean TSP each vertex $i$ always corresponds to one point $\point_i = (x_i , y_i)$ in the plane. 

The constraint model includes an {\tt alldifferent}$(L)$ constraint \cite{alldifferent} on the list $L$ of all variables,
that ensures that each node has exactly one incoming edge, as well a \circuit$(L)$ \cite{GlobalConstraintsCHIP,CaseauLaburthe,KayaHooker} constraint (sometimes called {\tt nocycle}) that avoids sub-tours, i.e., cycles of length less than \Npoints.

Caseau and Laburthe \cite{CaseauLaburthe} propose a simple but efficient propagation algorithm for {\tt circuit} constraint (nocycle) combined with new branching strategies based on the combination of first-fail and max-regret. %\mg{max regret, invece di regret}
First-fail selects first the variable with smallest domain because it has a higher probability of running out of elements and thus lead to failure. The regret of a variable is defined, instead, as the cost difference between its two best possible assignments, so the strategy is to first select those variables in which the regret is higher in order to avoid a significant increase in the cost of the solution.
They also propose to filter values based on the objective function. For this last purpose, they apply the assignment-based and the spanning tree relaxations. %\mg{relaxations}

Another filtering rule proposed for the \circuit\ constraint is that of Kaya and Hooker \cite{KayaHooker} based on graph separators theory.
Francis and Stuckey \cite{explainingCircuit} compare the effectiveness of different propagation algorithm for \circuit\ when adding explanation in the context of a lazy clause generation solver.

Pesant et al.~\cite{PesantGPR98} address the \ac{TSPTW}, in which cities must be visited within given temporal intervals, and exploit the \circuit\ constraint together with the minimum spanning tree relaxation. 
Focacci et al. \cite{FocacciLM_AMAI02,FocacciLM_Informs02} introduce hybrid approaches that merge \ac{CP} and Operations Research techniques, including: reduced costs filtering, use of the assignment problem and minimum spanning forest relaxation.

A filtering technique presented more recently by Benchimol et al.~\cite{BenchimolHRRR12} for the \ac{WCC} has been able to obtain, on medium dimension instances, results comparable with the solver Concorde.  In their work, they use a variety of techniques. In order to obtain an initial upper bound to be used for propagation, they first run the Lin-Kernighan-Helsgaun algorithm \cite{LinKernighan,Helsgaun00}.
Then the filtering technique uses the Held and Karp \cite{HeldKarp} scheme together with a Lagrangian relaxation to obtain the reduced costs of the edges and uses it to remove them. The algorithm is also able to identify the mandatory arcs that if removed would increase the current lower bound. In the experiments with asymmetric \acp{TSP}, they also use additive bounding \cite{AdditiveBoundingTSP} to combine both the 1-tree and the assignment problem relaxations.

Fages et al. \cite{SalesmanAndTree}, achieve significant improvement by casting the problem in CP(Graph) and through the introduction of improved search strategies (i.e. Last Conflict heuristic).
Fages and Lorca \cite{DBLP:journals/corr/abs-1206-3437} shown how properties of the reduced graphs associated to an Asymmetric TSPs can be used to improve the Minimum Spanning Tree relaxation. The nodes of the reduced graphs are the Strongly Connected Components of the original graph.

Isoart and R\'egin \cite{isoart2019integration} design a propagator based on the search %\mg{search, non research} 
of $k$-cutsets. The combination of this constraint with the \ac{WCC} constraint has resulted in a significant reduction in the computation time.

\section{Ongoing research}
Several algorithms have already been developed, some of them have recently been published in their first version \cite{bertagnonG20} 
while others are still being studied.

\subsection{The nocrossing constraint}
\begin{figure}[!htb]
\centering
\includegraphics[scale=1]{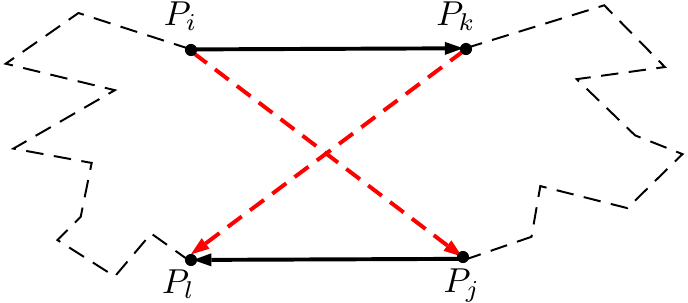}
\caption{Crossings elimination leads to a shorter route.} 
\label{fig1}
\end{figure}

A well-known finding in the literature (e.g., \cite{Arora_PTAS}) is that the optimal solution of Euclidean TSP does not have crossing edges (see Figure~\ref{fig1}).

\begin{property}\label{pty:nocrosses}
There is no crossing in the optimal solution for a Euclidean TSP. 
\end{property}

We can take advantage of Property~\ref{pty:nocrosses} to reduce the size of the search tree by eliminating branches presenting crosses. We propose the \nocrossing\ constraint that eliminates assignments leading to non-optimal solutions.  In the successor representation, it is defined as follows.

\begin{definition}
The $\nocrossing(i,\Next_i,j,\Next_j)$ constraint assures that segments $\segment{i}{\Next_i}$ and $\segment{j}{\Next_j}$ do not cross each other.
\label{def:nocrossing_definition}
\end{definition}

%\begin{equation}
%\nocrossing(i,\Next_i,j,\Next_j) = \left( \segment{\point_i}{\point_{\Next_{i}}} \cap \segment{\point_j}{\point_{\Next_{j}}} \right)
%\subset \{\point_i,\point_j\}.
%\label{eq:nocrossing_definition}
%\end{equation}

The \nocrossing\ constraint is a binary constraint since $i$ and $j$ are ground values at the time when it is imposed. 
Suppose we have $n$ nodes, to avoid crossings $n(n-1)/2$ constraints are introduced, one constraint for each pair of nodes of the graph.
The \nocrossing\ constraint can be implemented thorough a pair of propagators: one propagates changes in the domain of the variable $Next_i$ on the domain of the variable $Next_j$, while the other propagator propagates changes on the other way. 

The \nocrossing\ constraint could be implemented in a naive manner, for example using the table constraint \cite{tableConstraintNengFa} or the {\tt propia} library \cite{propia}. A table constraint consists of a table (usually a list of tuples) of values that the involved variables must, or must not, assume. However this implementation would be inefficient, because the constraint wakes up most of the time without being able to carry out any propagation. Moreover, with the table constraint one should initially compute large tables, containing, for all pairs of edges in the graph, if they cross or not. Propagating such constraint would have the usual cost of arc-consistency propagation for a single constraint of $O(d^2)$ (if $d$ is the size of the domains) in each activation of the constraint.

\begin{figure*}[tb]
\centering
\sidesubfloat[]{%
	 \includegraphics[width=0.44\linewidth]{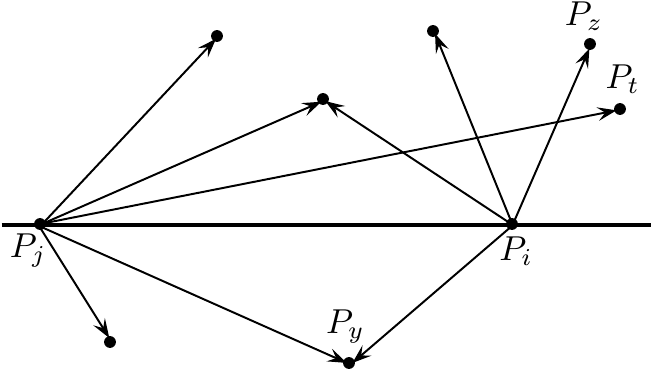}\label{fig:crossing_example}
}\hspace{1mm}%
\sidesubfloat[]{%
	  \includegraphics[width=0.44\linewidth]{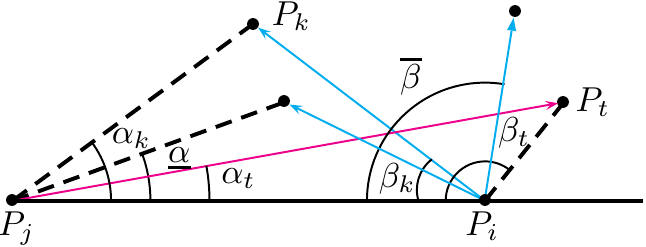}\label{fig:phase2}
}
\caption{An arrow from $\point_x$ to $\point_y$ means that $y \in \Dom{\Next_x}$. Dashed lines are plotted to show the angles.}
\end{figure*}

From the definition of arc-consistency and Definition~\ref{def:nocrossing_definition},
a value $v$ can be removed from $\Dom{\Next_j}$ only if the segment \segment{\point_j}{\point_v} intersects all possible segments originating from $\point_i$ (e.g. segment \segment{\point_j}{\point_t} in Figure~\ref{fig:crossing_example}).

%\begin{figure}
%\centering
%\includegraphics[scale=1]{points_sep}
%\caption{An arrow from $\point_x$ to $\point_y$ means that $y \in \Dom{\Next_x}$. }
%\label{fig:crossing_example}
%\end{figure}

We define a {\em possible segment} from $\point_i$ a segment $\segment{\point_i}{\point_j}$ such that $j \in \Dom{\Next_i}$. 
%\mg{questo admissible non mi piace moltissimo; da` l'idea di una proprieta` un po' piu` strutturale, che non dipende solo dal fatto che adesso e` nel dominio.
%Forse qualcosa come "current segment" o "currently considered segment", ...}
We denote with $\StraightLine{\point_i}{\point_j}$ the (infinite straight) line passing through points $\point_i$ and $\point_j$, and
with $\Angle{\point_i\point_j\point_k}$ the counterclockwise angle formed by the segments $\segment{\point_i}{\point_j}$ and $\segment{\point_j}{\point_k}$ with vertex in $\point_j$ from $\point_i$ to $\point_k$. 

We have identified the following necessary and sufficient conditions for pruning and we have exploited them within our propagation algorithm.

\begin{theorem} [Necessary condition]
%A segment $\segment{\point_j}{\point_t}$ originating from $\point_j$ and reaching an element $t \in \Dom{\Next_j}$ can cross all segments originating from $\point_i$ only if all segments originating from $\point_j$ lie on the same half-plane with respect to the line \StraightLine{\point_i}{\point_j}.
Let $\segment{\point_j}{\point_t}$ a possible segment from $\point_j$. If $\segment{\point_j}{\point_t}$ crosses all possible segments from $\point_i$ then all segments originating from $\point_i$ must lie on the same half-plane with respect to the line \StraightLine{\point_i}{\point_j}.
\end{theorem}

The propagator is suspended waiting that all elements in the domain of $\Next_i$ lie on the same half-plane. We select one element in each half-plane and the propagator suspends waiting that one of the two elements is removed from $\Dom{\Next_i}$.

\begin{theorem} [Necessary condition]
%A segment $\segment{\point_j}{\point_t}$ originating from $\point_j$ and reaching an element $t \in \Dom{\Next_j}$
%can cross all segments $\segment{\point_i}{\point_q}$ originating from $\point_i$ only if
%$\alpha_t \leq \minAlpha$,
%where 
%$\minAlpha = \min \{ \alpha_q \mid q \in \Dom{\Next_i}\}$ and $\alpha_k=\Angle{\point_i\point_j\point_k}$ (see figure~\ref{fig:phase2}).

Let $\segment{\point_j}{\point_t}$ a possible segment from $\point_j$. If $\segment{\point_j}{\point_t}$ crosses all possible segments from $\point_i$ then $\alpha_t \leq \minAlpha$,
where 
$\minAlpha = \min \{ \alpha_q \mid q \in \Dom{\Next_i}\}$ and $\alpha_k=\Angle{\point_i\point_j\point_k}$ (see Figure~\ref{fig:phase2}).
\end{theorem}

\begin{theorem} [Sufficient condition]
Suppose there exists a possible segment $\segment{\point_j}{\point_t}$ from $\point_j$ such that $\alpha_t \leq \minAlpha$.
If $\segment{\point_j}{\point_t}$ crosses all possible segments from $\point_i$ then $\beta_t > \maxBeta$, where $\maxBeta = \max\{\beta_q \mid q \in \Dom{\Next_i} \}$ and $\beta_k = \Angle{\point_k\point_i\point_j}$ (see Figure~\ref{fig:phase2}).
\end{theorem}

%\begin{figure}
%\centering
%\includegraphics[scale=1]{phase2_sufficient}
%\caption{%An arrow from $\point_x$ to $\point_y$ means that $y \in \Dom{\Next_x}$. 
%Dashed lines are plotted to show the angles.\label{fig:phase2}}
%\end{figure}

Thanks to these theorems, the \nocrossing\ propagator can be implemented with ${O(d)}$ complexity (if $d$ is the size of the domains) per each activation (to be compared to the $O(d^2)$ of a naive propagator). 

\begin{figure}[tb]
\centering
\sidesubfloat[]{%
	 \includegraphics[width=0.44\linewidth]{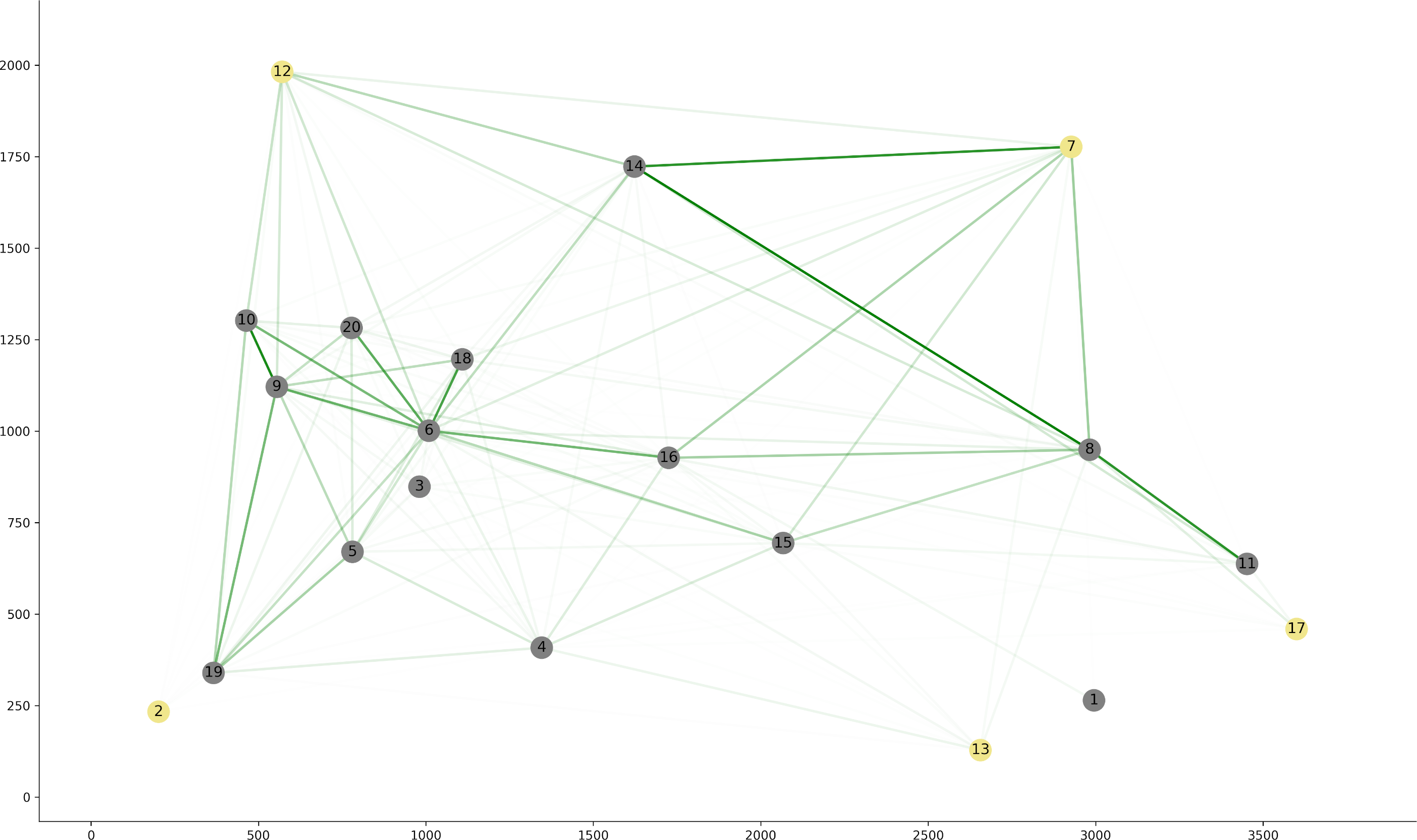}
}\hspace{1mm}%
\sidesubfloat[]{%
	  \includegraphics[width=0.44\linewidth]{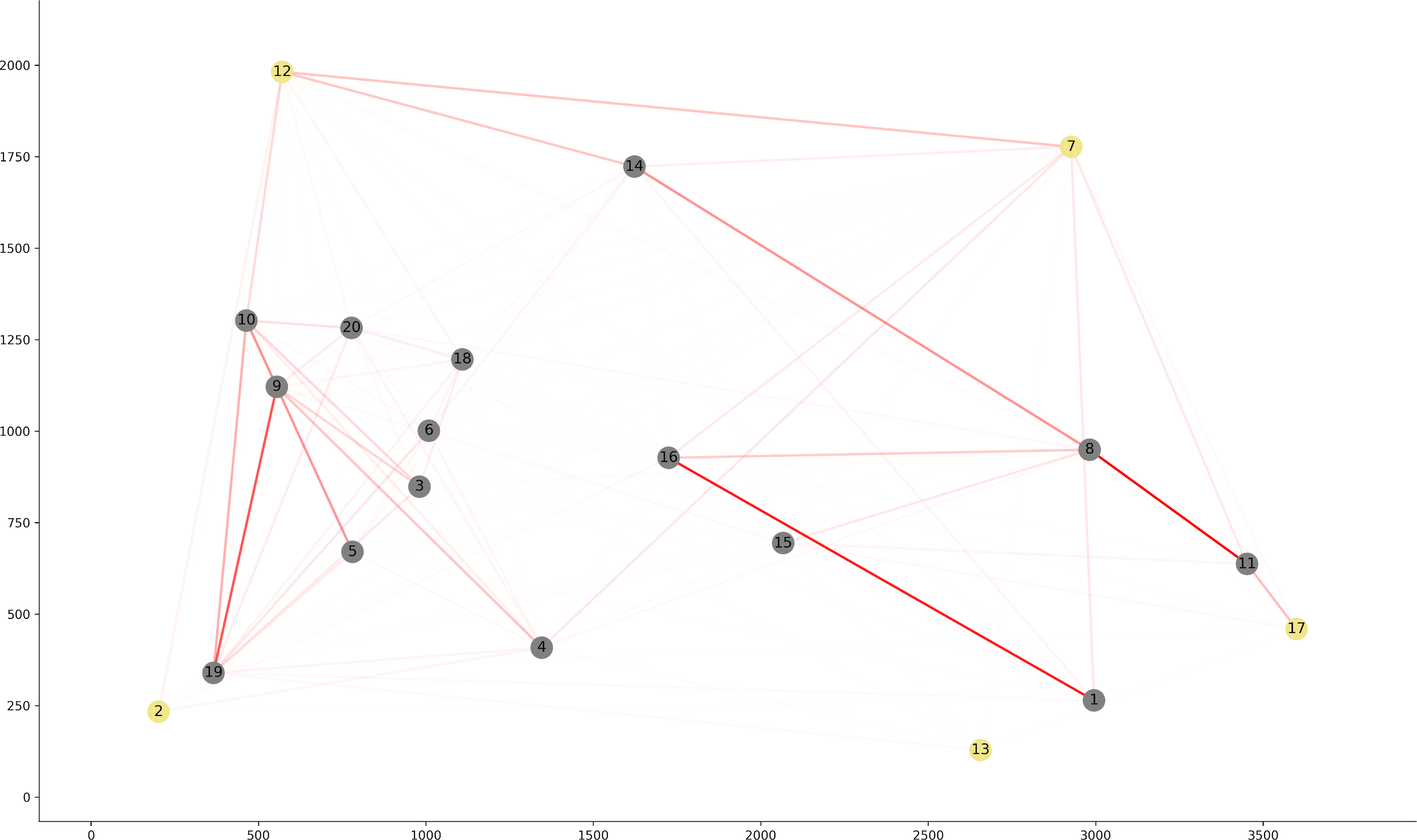}
}\hspace{1mm}%
\sidesubfloat[]{%
	  \includegraphics[width=0.44\linewidth]{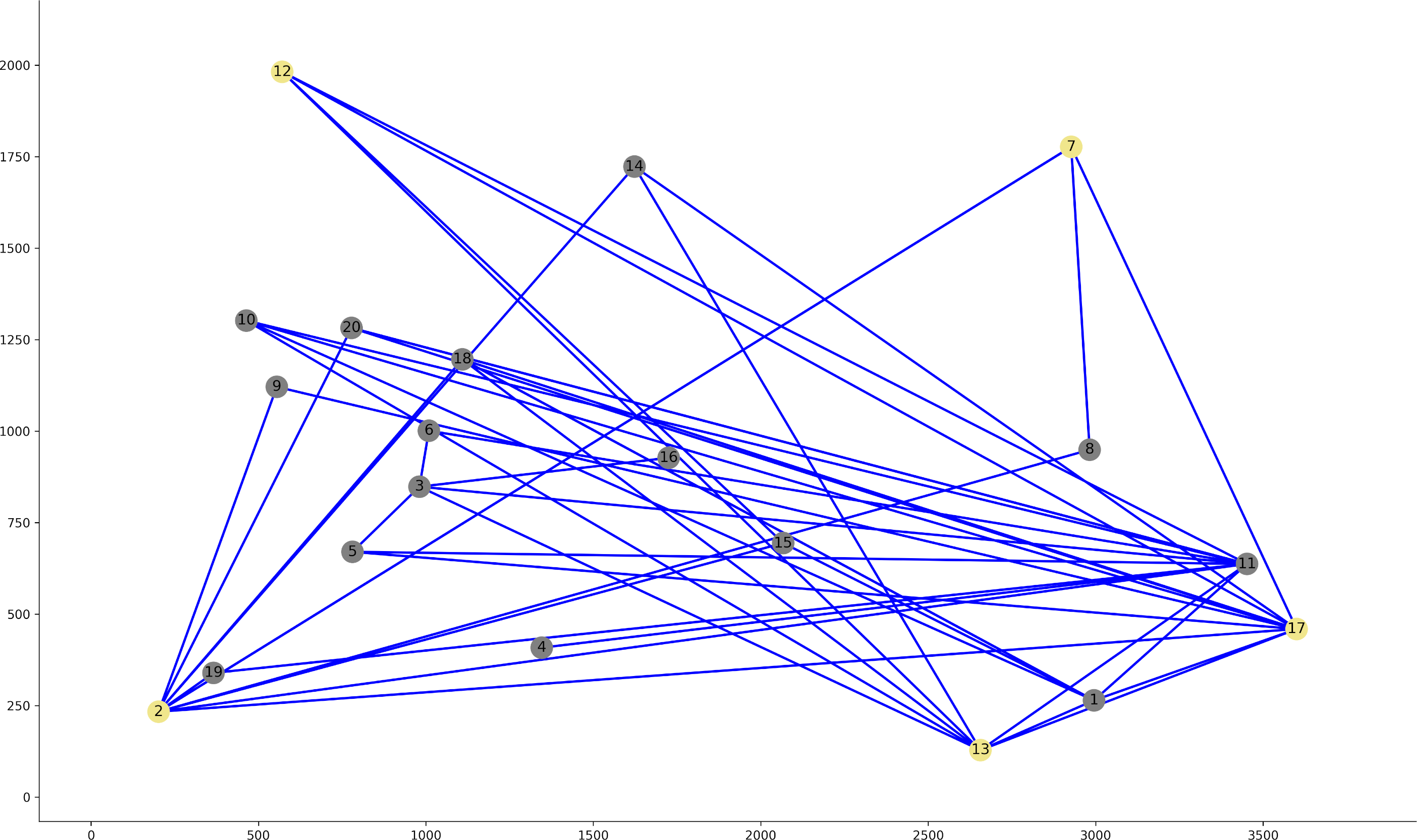}
}
\caption{A graphical representation of the \nocrossing\ constraint while solving an instance with 20 nodes. The darker the color, the higher is the value for the following indicators: (a) number of value deletion; (b) number of failures; (c) instances of \nocrossing\ that have not performed pruning (binary value: true, false).}
\label{fig:statistics}
\end{figure}

A study we are conducting to evaluate the actual performance of the \nocrossing\ constraint is shown in Figure~\ref{fig:statistics}.

%\begin{figure}[tb]
%\centering
%\includegraphics[scale=0.27]{figure/kroD100.png}
%\caption{A graphical representation of the \nocrossing\ constraint while solving an instance with 20 nodes. The darker the color, the higher is the value for the following indicators: green (upper left) number of value deletion; red (upper right) number of failures; orange (lower left) the ratio of value deletions over the number of activation of the constraint; blue (lower right) instances of \nocrossing\ that have not performed pruning (binary value: true, false).
%\mg{Come mai ci sono dei segmenti che sono sia arancioni sia blu? Per esempio il 7-17.}
%} 
%\label{fig:statistics}
%\end{figure}

\subsection{The clockwise constraint}
\label{sec:convex}
Further pruning can be obtained by means of the convex hull of the set of points. We recall the definition of convex hull.
%\ab{qui, da qualche parte, dovrei citare anche la figura 4a}
\begin{definition}
Let $S$ be a set of points, the convex hull of $S$ is defined as the smallest convex set containing $S$. The convex hull is therefore the smallest convex polygon that includes all the points of the set $S$.
\end{definition}

The following property resulting from Property~\ref{pty:nocrosses} can also be exploited to reduce the space of possible solutions.

\begin{property}\label{property:hull}
Let $k$ of the $n$ points in the Euclidean TSP be vertices on the boundary of the convex hull. Then the order in which these $k$ points appear in the optimum traveling salesman tour must be the same as the order in which these same points appear on the boundary of 
the convex hull.
\end{property}

\begin{figure*}[tb]
\centering
\sidesubfloat[]{%
	 \includegraphics[width=0.30\linewidth]{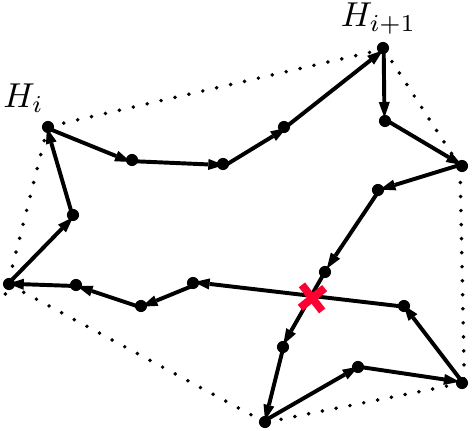}\label{fig:hull_cross}
}\hspace{15mm}%
\sidesubfloat[]{%
	  \includegraphics[width=0.28\linewidth]{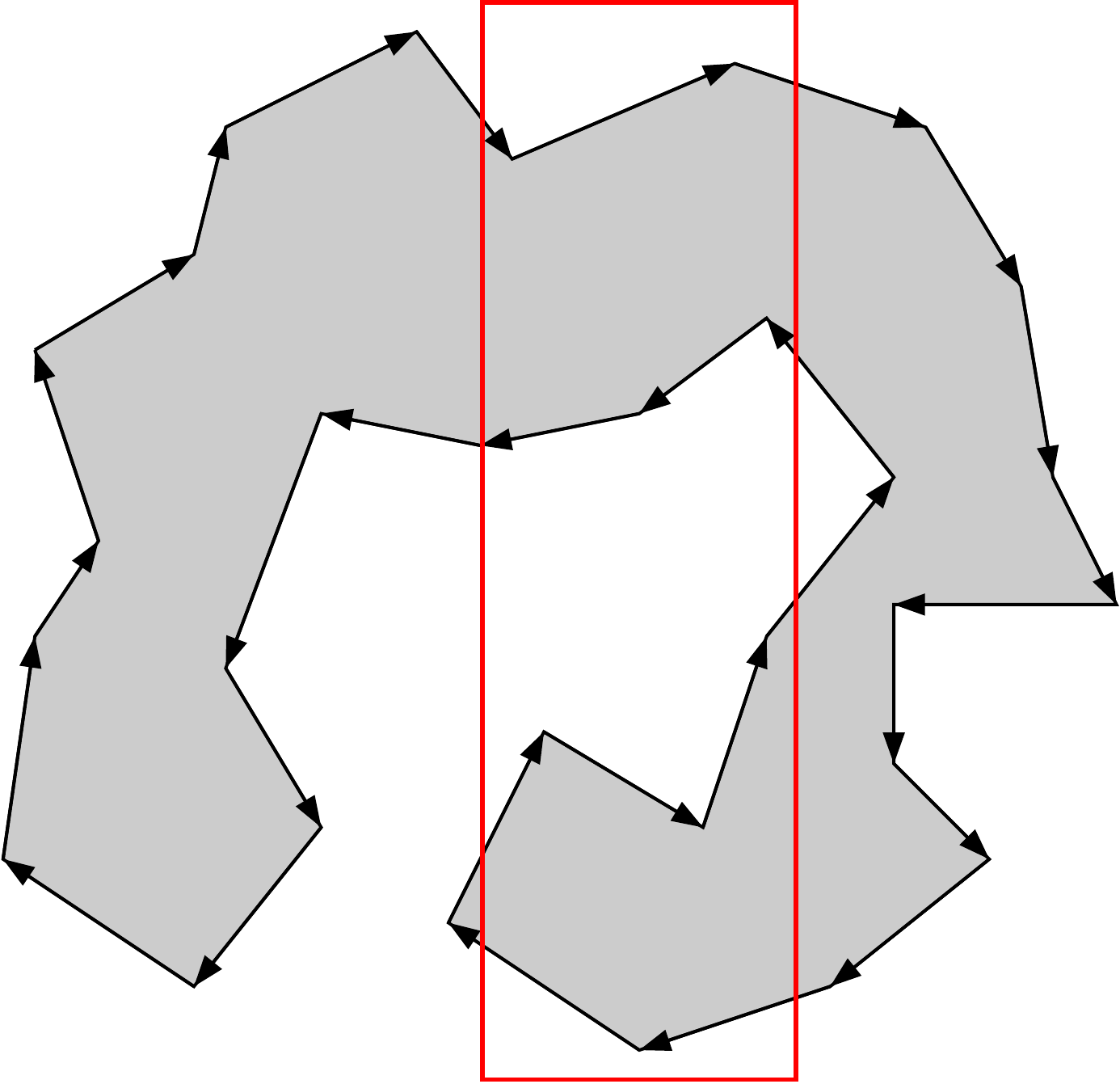}\label{fig:tsp}
}
\caption{(a) Visiting vertices of the perimeter of the convex hull not in order generates a cross in the solution of the TSP, therefore that solution cannot be optimal. (b) Imagine to cut the optimal TSP with two vertical lines, the convex hull reasoning can be extended to the borders (i.e., the parts of the circuit inside the stripe).}
\label{clockwise}
\end{figure*}

From Property~\ref{property:hull} follows that the optimal solution of the TSP is is a simple polygon, and it divides the plane into exactly two areas: an {\em internal} and an
{\em external} area.
Please observe that the points on the border of the convex hull do not necessarily appear consecutively in the solution, an example is reported in Figure~\ref{fig:hull_cross}.

%An example representing the situation described in Property~\ref{property:hull} is shown in Figure~\ref{fig7}. In addition to \nocrossing\ constraint we propose the \clockwise\ constraint based on Property~\ref{property:hull}.
%
%\begin{definition}
%The $\clockwise\ (L)$ constraint ensures that the circuit described by the list $L$ has vertices of the boundary of the convex hull visited clockwise.
%\end{definition}

%\begin{figure}
%\centering
%\includegraphics[scale=3]{figure/fig14.eps}
%\caption{Visiting vertices of the perimeter of the convex hull not in order generates a cross in the solution of the TSP, therefore that solution cannot be optimal.} 
%\label{fig7}
%\end{figure}

Let $V= \lbrace 1,2,...,n \rbrace$ be the set of vertices, where $n$ is the number of vertices of the graph, and let $\Hull = \lbrace (h_{1},h_{2},...,h_{k}) \vert h_{i} \in V, i=1,..,k \rbrace$ be the list of vertices that lies on the boundary of the convex hull ordered clockwise. The following is a summary of three different ways we proposed to exploit the information about the convex hull for constraint propagation.

\begin{figure*}[tb]
\centering
\sidesubfloat[]{%
	 \includegraphics[width=0.25\linewidth]{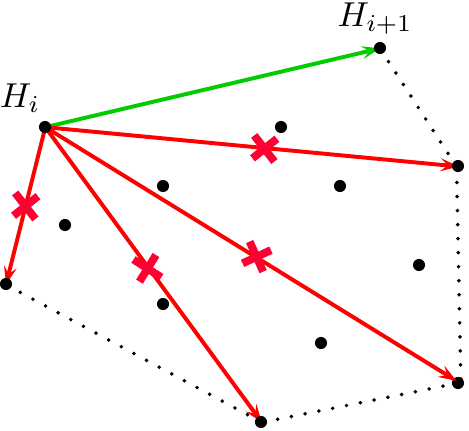}\label{fig:hull1}
}\hspace{1mm}%
\sidesubfloat[]{%
	  \includegraphics[width=0.28\linewidth]{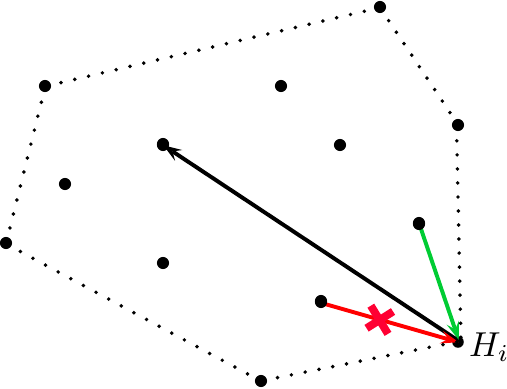}\label{fig:hull2}
}\hspace{1mm}%
\sidesubfloat[]{%
	  \includegraphics[width=0.30\linewidth]{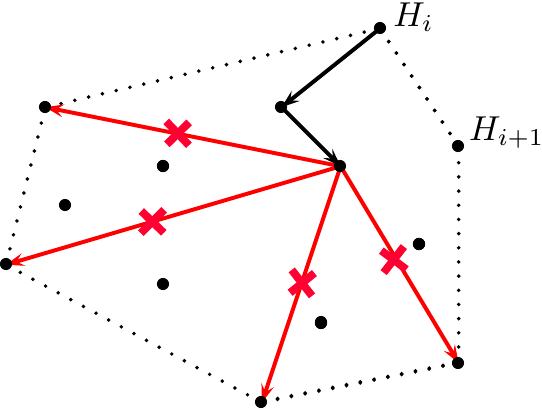}\label{fig:hull3}
}
\label{fig:hulls}
\caption{Depiction of three different ways to exploit the information about the convex hull for constraint propagation as implemented in the \clockwise\ constraint. The boundary of the convex hull is represented with dotted lines. (a) The successor of a convex hull vertex cannot be another vertices on the boundary of the convex hull except for the one that immediately follows it. (b)  In order to visit nodes in a clockwise order, the angle between the incoming edge and the outgoing edge of a convex hull vertex cannot be positive (it must be between $-\pi$ and 0). (c) On each paths originating from a convex hull vertex $\Hull_i$ the first convex hull vertex different from $\Hull_i$ is the one immediately following it, that is $\Hull_{i+1}$.}
\end{figure*}

The simplest way to satisfy the Property~\ref{property:hull} is to apply the following relation:
\begin{equation}
\forall  i \in \Hull,  \:  \Dom{\Next_{i}} \cap \lbrace h_{j}  | j = 1,...k,  j \neq (i+1) \rbrace = \emptyset
\label{eq:hullpruning}
\end{equation}
%\ab{la dicitura X\_i non mi convince}
Equation~\ref{eq:hullpruning} states that it is possible to eliminate from the domain of each point in $\Hull$ all the other points on the boundary of the convex hull except the next in the sequence (see Figure~\ref{fig:hull1}).

Further propagation can be carried out when the domain associated with a point on the boundary of the convex hull becomes ground.
Let $\Hull_{i}$ be the index of the point on the boundary of the convex hull whose domain has become ground and $P$ its value, i.e., in the current assignment there is the $\segment{\Hull_{i}}{P}$ segment. If we choose to go through the points on the boundary of the convex hull in a clockwise direction, in order to satisfy Property~\ref{property:hull}, no points situated on the left of the $\segment{\Hull_{i}}{P}$ segment can have $\Hull_{i}$ as its successor (see Figure~\ref{fig:hull2}) 
%\ab{qui non è ben chiaro il verso di percorrenza della hull che va fissato a priori}

The third way is to impose that each path originating from a convex hull vertex cannot reach any convex hull vertex except for the one immediately following it. Implementation is inspired by the \circuit\ constraint \cite{CaseauLaburthe}, but performs more powerful pruning (see Figure~\ref{fig:hull3}). It is clear that this third propagation also implies the first one.

%\begin{figure}
%\centering
%\includegraphics[scale=1]{figure/hull1.pdf}
%\caption{The successor of a convex hull vertex cannot be another member of $\HullSet$ except for the one that immediately follows it.} 
%\label{fig:hull1}
%\end{figure}
%
%\begin{figure}
%\centering
%\includegraphics[scale=1]{figure/hull2.pdf}
%\caption{In order to visit nodes in a clockwise order, the angle between the incoming edge and the outgoing edge of a convex hull vertex cannot be positive (it must be between $-\pi$ and 0) or, stated otherwise, it must correspond to a right turn.} 
%\label{fig:hull2}
%\end{figure}
%
%\begin{figure}
%\centering
%\includegraphics[scale=1]{figure/hull3.pdf}
%\caption{Each path originating from a convex hull vertex cannot reach any convex hull vertex except for the one immediately following it. Implementation is inspired by the \circuit\ constraint (\cite{CaseauLaburthe}), but performs more powerful pruning.} 
%\label{fig:hull3}
%\end{figure}

%\begin{figure}[!ht]
%\minipage{0.27\textwidth}
%  \includegraphics[width=\linewidth]{figure/hull1.pdf}
%  \caption{1. Perimeter Order}\label{fig:hull1}
%\endminipage\hfill
%\minipage{0.33\textwidth}
%  \includegraphics[width=\linewidth]{figure/hull2.pdf}
%  \caption{2. Incoming Clockwise}\label{fig:hull2}
%\endminipage\hfill
%\minipage{0.34\textwidth}%
%  \includegraphics[width=\linewidth]{figure/hull3.pdf}
%  \caption{3. Hull Path}\label{fig:hull3}
%\endminipage
%\end{figure}
%\end{block}

%Extended convex hull reasoning
The applicability of the three convex hull propagators presented can also be extended to points that lie in the interior of the hull. Once a partial path has been defined, consider the polygon formed by such path plus a segment connecting its extremes. We can apply the propagators to the convex hull of the points inside that polygon (see Figure~\ref{fig:tsp}).

We propose the \clockwise\ constraint that implements all the propagation described in this section.

\section{Preliminary Result}
In order to assess the effectiveness of the proposed algorithms, we devised a series of experiments based on randomly-generated TSPs and taken from structured instances (e.g. TSPLIB  \cite{tsplib}).

We implemented all algorithms using the \ECLiPSe\ CLP language \cite{ECLiPSe}. We started comparing our algorithms with a simple model for the successor representation (that includes \alldifferent\ and \circuit\ constraints), such a constraint model is named {\tt \eclipseBase} in the following. Then in order to show that the pruning we provide is not subsumed by that of state of the art techniques, we implemented in \ECLiPSe, in the successor representation, also the Held and Karp bound with pruning based on reduced and marginal costs, as proposed by Benchimol et al. \cite{BenchimolHRRR12} (shown with {\tt \loro} in the following).
The constraint model, named {\tt \noi}, includes the \nocrossing\ constraint on all pairs of vertices, together with the  \clockwise\ constraint that implements the propagation described in Section~\ref{sec:convex}. The {\tt \noi} model also includes bounds with the Lin-Kernighan-Helsgaun \cite{Helsgaun00}) as included in {\tt \loro}. %\mg{Detto cosi` sembra che siamo unfair: nel nostro modello c'\`e il symmetry breaking e nel loro no.}

In Figure~\ref{fig:graphs-thesis} we present cactus plots that compare the impact of our filtering algorithms with the constraints already predefined in $ECL^{i}PS^{e}$ system. In the $x$-axis we report the number of optimally solved instances, and in the $y$-axis the solving time in seconds. Fixed a certain amount of time, Figure~\ref{fig:graphs-thesis} clearly illustrates that our filtering algorithms, especially when used simultaneously, significantly increase the number of instances solved.

In Figure~\ref{fig:graphs-aaai20} we present cactus plot that compare the impact of our filtering algorithm with the pruning proposed by Benchimol et al. \cite{BenchimolHRRR12}.  The addition of the filtering on geometric properties improves the runtime.

Admittedly, the instances we were able to solve are not as large as those addressed by state of the art techniques in CP \cite{BenchimolHRRR12,SalesmanAndTree,DBLP:journals/corr/abs-1206-3437}.
This could be due to several factors. 
Our implementation is based on declarative languages that have the advantage of separating problem definition (as much as possible) from implementation details. On the other hand, their performance can be significantly slower than imperative languages, not only due to interpretation vs compilation schemes, but also to the availability of efficient data structures and search techniques.

Although the results achieved so far are a solid starting point, we are still working on improving the performance and making our algorithms scalable to solve larger instances. 

\newcommand{\mycircle}[1]{\tikz{\filldraw[draw=#1,fill=#1] (0,0) circle [radius=0.3em];}}
\tikzset{
    square/.style={
    		line width=0.3pt,
        	postaction={
        		decoration={
          		markings,
          		mark=between positions 0 and 1 step 0.07
               	with {\filldraw[solid] (-1pt,1pt) -- (1pt,1pt) -- (1pt,-1pt) -- (-1pt,-1pt); },
        		},
        		decorate,
     	},
     },
     triangle/.style={
    		line width=0.3pt,
        	postaction={
        		decoration={
          		markings,
          		mark=between positions 0 and 1 step 0.07
               	with { \filldraw[solid] (-1.5pt,-1.5pt) -- (0pt,1.5pt) -- (1.5pt,-1.5pt); },
        		},
        		decorate,
     	},
     },
     diamond/.style={
    		line width=0.3pt,
        	postaction={
        		decoration={
          		markings,
          		mark=between positions 0 and 1 step 0.07
               	with { \filldraw[solid] (-1.5pt,0pt) -- (0pt,1.5pt) -- (1.5pt,0pt) -- (0pt,-1.5pt);  },
        		},
        		decorate,
     	},
     },
}

\floatsetup[figure]{style=plain,subcapbesideposition=top, capbesideposition={top,right}}
\thisfloatsetup{subfloatrowsep=none}
\begin{figure*}[t]
\centering

\sidesubfloat[]{%
		\begin{tikzpicture}
		\begin{axis}[
		xmin=1, xmax=68, ymin=0, ymax=1170,
		minor x tick num= 1,
		minor y tick num= 1,
		axis x line=bottom,
		axis y line=left,
		width = 0.42\columnwidth,
		xlabel={\# solved instances}, ylabel={time [s]}, title={},
		axis on top=true,
		legend pos = north west,
		legend style={font=\fontsize{7}{5}\selectfont},
		label style={font=\footnotesize},
		tick label style={font=\footnotesize},
		tick label style={/pgf/number format/assume math mode=true},
		y label style={at={(axis description cs:-0.14,.5)},anchor=south},
		]
		\addplot [cyan, diamond, forget plot, line width=1pt] table [x=X, y=Y, col sep=semicolon] {plotdata/noconstr_20.csv};
		\addplot [magenta, square, forget plot, line width=1pt] table [x=X, y=Y, col sep=semicolon] {plotdata/nocross_20.csv};
		\addplot [black, triangle, forget plot, line width=1pt] table [x=X, y=Y, col sep=semicolon] {plotdata/constr_20.csv};
		%\legend{{CLP(FD)},{CLP(FD) + \nocrossing\ },{CLP(FD) + \clockwise\ },{CLP(FD) + \nocrossing\ \,+ \clockwise\ }}
		\end{axis}
		\end{tikzpicture}
		\label{graph:20nodes}
}\hspace{1mm}%
\sidesubfloat[]{%
		\begin{tikzpicture}
		\begin{axis}[
		xmin=1, xmax=50, ymin=0, ymax=1170,
		minor x tick num= 1,
		minor y tick num= 1,
		axis x line=bottom,
		axis y line=left,
		width = 0.42\columnwidth,
		xlabel={\# solved instances}, ylabel={time [s]}, title={},
		axis on top=true,
		legend pos = south west,
		legend style={font=\fontsize{6.4}{5}\selectfont, draw=none, fill=none, at={(0.80,0.12)}},
		legend style={cells={align=left}},
		legend image post style={scale=0.8},  
		label style={font=\footnotesize},
		tick label style={font=\footnotesize},
		y label style={at={(axis description cs:-0.14,.5)},anchor=south}
		]
		\addplot [cyan, diamond, forget plot, line width=1pt] table [x=X, y=Y, col sep=semicolon] {plotdata/noconstr_24.csv};
		\addlegendentry{\eclipseBase}
		\addplot [magenta, square, forget plot, line width=1pt] table [x=X, y=Y, col sep=semicolon] {plotdata/nocross_24.csv};
		\addlegendentry{\eclipseBase\ + \\
										 \nocrossing\  
									   }
		\addplot [black, triangle, forget plot, line width=1pt] table [x=X, y=Y, col sep=semicolon] {plotdata/constr_24.csv};
						\addlegendentry{\eclipseBase\ + \\
						\nocrossing\ \,+ \\
						\clockwise\  
									   }
		%\legend{{CLP(FD)},{CLP(FD) + \nocrossing\ },{CLP(FD) + \clockwise\ },{CLP(FD) + \nocrossing\ \,+ \clockwise\ }}

		\addlegendimage{cyan, only marks, mark=diamond*}
		\addlegendimage{magenta, only marks, mark=square*}
		\addlegendimage{black, only marks, mark=triangle*}

		\end{axis}
		\end{tikzpicture}
		\label{graph:24nodes}
}
\caption{Cactus plot of filtering algorithms which we run over 68 randomly-generated Euclidean TSP instances with 20 nodes (a) and 24 nodes (b). Time limit was set to 1200 seconds.}
\label{fig:graphs-thesis}
\end{figure*}
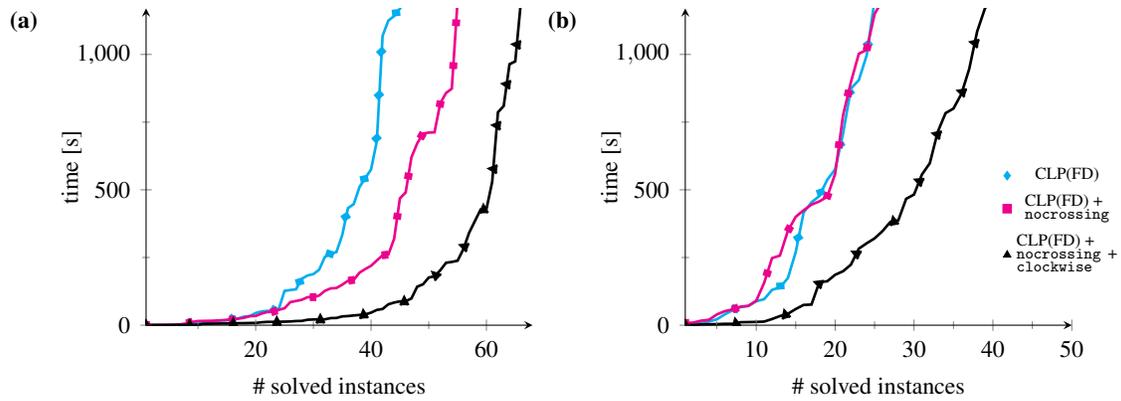

\begin{figure*}[t]
\centering

\sidesubfloat[]{%
		\begin{tikzpicture}[baseline]
		\begin{axis}[
		ymode=log,
       	log basis y={10},
		xmin=20, xmax=50, ymin=0.2, ymax=100,
		axis x line=bottom,
		axis y line=left,
		width = 0.42\textwidth,
		xlabel={\# nodes}, ylabel={$\log_{10}$ time [s]}, title={},
		axis on top=true,
		legend pos = south west,
		legend style={font=\fontsize{6.4}{5}\selectfont, draw=none, fill=none, at={(1.02,0.00)}},
		legend image post style={scale=0.8},  
		label style={font=\footnotesize},
		tick label style={font=\footnotesize},
		y label style={at={(axis description cs:-0.11,.5)},anchor=south},
		]
				
%		\addplot [cyan, diamond, forget plot, dash pattern=on 1pt off 1pt] table [x=x, y=y0, col sep=comma] {plotdata/ntime_maxregret.csv};
%		\addplot [cyan, diamond, line width=1pt, forget plot, dash pattern=on 1pt off 1pt] table [x=x, y=y1, col sep=comma] {plotdata/ntime_maxregret.csv};
%		
%		\addplot [magenta, square, forget plot, dash pattern=on 1pt off 1pt] table [x=x, y=y0, col sep=comma] {plotdata/ntime_lkh_lcfirst.csv};
%		\addplot [magenta, square, line width=1pt, forget plot, dash pattern=on 1pt off 1pt] table [x=x, y=y1, col sep=comma] {plotdata/ntime_lkh_lcfirst.csv};
%		
%		\addplot [black, triangle, forget plot, dash pattern=on 1pt off 1pt] table [x=x, y=y0, col sep=comma] {plotdata/ntime_lkh_maxregret.csv};
%		\addplot [black, triangle, line width=1pt, forget plot, dash pattern=on 1pt off 1pt] table [x=x, y=y1, col sep=comma] {plotdata/ntime_lkh_maxregret.csv};
%		
%		\addplot [cyan, diamond, forget plot] table [x=x, y=y2, col sep=comma] {plotdata/ntime_maxregret.csv};
%		\addplot [cyan, diamond, line width=1pt, forget plot] table [x=x, y=y3, col sep=comma] {plotdata/ntime_maxregret.csv};
		
%		\addplot [magenta, square, forget plot] table [x=x, y=y2, col sep=comma] {plotdata/ntime_lkh_lcfirst.csv};
%		\addplot [magenta, square, line width=1pt, forget plot] table [x=x, y=y3, col sep=comma] {plotdata/ntime_lkh_lcfirst.csv};
		
		\addplot [black, triangle, line width=1pt, forget plot] table [x=x, y=y2, col sep=comma] {plotdata/ntime_lkh_maxregret.csv};
		\addplot [magenta, square, line width=1pt, forget plot] table [x=x, y=y3, col sep=comma] {plotdata/ntime_lkh_maxregret.csv};
		
		\end{axis}
		\end{tikzpicture}
		\label{fig:time_plot}
}\hspace{1mm}%
\sidesubfloat[]{%
		\begin{tikzpicture}[baseline]
		\begin{axis}[
		xmin=600, xmax=970, ymin=0, ymax=1900,
		axis x line=bottom,
		axis y line=left,
		width = 0.42\textwidth,
		ytick={600,1200,1800},
		xlabel={\# solved instances}, ylabel={time [s]}, title={},
		axis on top=true,
		legend pos = south west,
		legend style={font=\fontsize{6.4}{5}\selectfont, draw=none, fill=none, at={(0.30,0.42)}},
		legend image post style={scale=0.8},  
		label style={font=\footnotesize},
		tick label style={font=\footnotesize},
		y label style={at={(axis description cs:-0.20,.5)},anchor=south},
		]

%		\addplot [cyan, diamond, forget plot, dash pattern=on 1pt off 1pt] table [x="x", y="y", col sep=comma] {plotdata/cactus_maxregret_0_spline.csv};
%		\addplot [cyan, diamond, line width=1pt, forget plot, dash pattern=on 1pt off 1pt] table [x="x", y="y", col sep=comma] {plotdata/cactus_maxregret_1_spline.csv};
%		
%		\addplot [magenta, square, forget plot, dash pattern=on 1pt off 1pt] table [x="x", y="y", col sep=comma] {plotdata/cactus_lkh_lcfirst_0_spline.csv};
%		\addplot [magenta, square, line width=1pt, forget plot, dash pattern=on 1pt off 1pt] table [x="x", y="y", col sep=comma] {plotdata/cactus_lkh_lcfirst_1_spline.csv};
%		
%		\addplot [black, triangle, forget plot, dash pattern=on 1pt off 1pt] table [x="x", y="y", col sep=comma] {plotdata/cactus_lkh_maxregret_0_spline.csv};
%		\addplot [black, triangle, line width=1pt, forget plot, dash pattern=on 1pt off 1pt] table [x="x", y="y", col sep=comma] {plotdata/cactus_lkh_maxregret_1_spline.csv};
%		
%		\addplot [cyan, diamond, forget plot] table [x="x", y="y", col sep=comma] {plotdata/cactus_maxregret_2_spline.csv};
%		\addplot [cyan, diamond, line width=1pt, forget plot] table [x="x", y="y", col sep=comma] {plotdata/cactus_maxregret_3_spline.csv};
		
%		\addplot [magenta, square, forget plot] table [x="x", y="y", col sep=comma] {plotdata/cactus_lkh_lcfirst_2_spline.csv};
%		\addplot [magenta, square, line width=1pt, forget plot] table [x="x", y="y", col sep=comma] {plotdata/cactus_lkh_lcfirst_3_spline.csv};
		
		\addplot [black, triangle, line width=1pt, forget plot] table [x="x", y="y", col sep=comma] {plotdata/cactus_lkh_maxregret_2_spline.csv};
		\addplot [magenta, square, line width=1pt, forget plot] table [x="x", y="y", col sep=comma] {plotdata/cactus_lkh_maxregret_3_spline.csv};
		
%		\addlegendimage{cyan, only marks, mark=diamond*}
		\addlegendimage{black, mark=triangle*}
		\addlegendimage{magenta, mark=square*}
%		\addlegendimage{no markers, dash pattern=on 1pt off 1pt}
%		\addlegendimage{no markers, dash pattern=on 1pt off 1pt, line width=1pt}
%		\addlegendimage{no markers}
%		\addlegendimage{no markers, line width=1pt}
		
		\legend{{\texttt{\loro}},{\texttt{\noipiuloro}}}
		
		\end{axis}
		\end{tikzpicture}
		\label{fig:cactus_plot}
}
\caption{Experimental results on randomly-generated Euclidean TSP instances from 20 to 50 nodes in steps of 2. For each size we generated 60 instances (30 uniform and 30 clustered).
Time limit was set to 1800 seconds. (a) Average solving time of filtering algorithms varying the size of the instances. Each point is the average solving time of 60 instances. (b) Cactus plot showing the number of solved instances varying the solving time.}
\label{fig:graphs-aaai20}
\end{figure*}
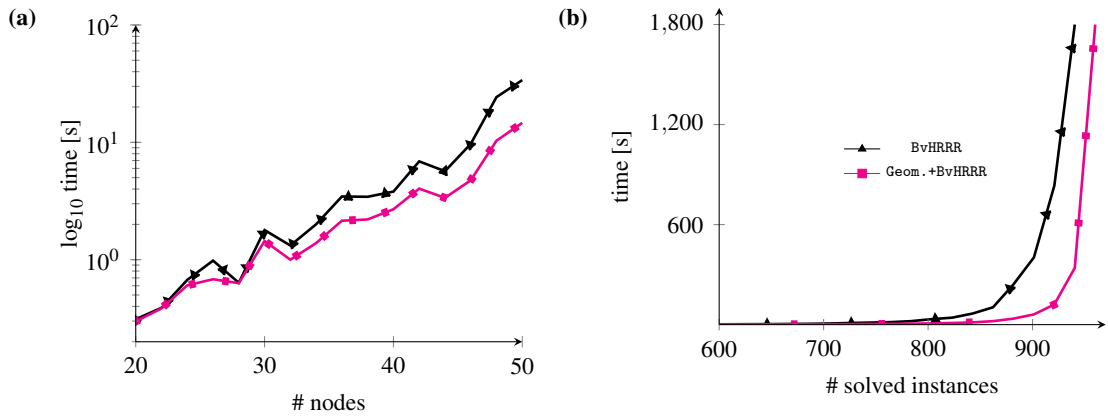

\section{Conclusion and Future Works}
In this paper, we proposed to exploit geometric information while solving route planning problems in order to have additional pruning
with respect to the techniques already available in CP. 

The pruning of the \ac{WCC}, proposed by Benchimol et al. \cite{BenchimolHRRR12}, is orthogonal with respect to that of the \nocrossing\ and \clockwise\ constraints. The quality of the propagation carried out by the \ac{WCC} is proportional to the quality of the current upper bound on the cost of the solution of the TSP, while our propagator can delete values from the domains even if no bound is known yet.

Despite the preliminary results that we have achieved and presented in this paper are good, our approaches are still not competitive with Concorde especially for large instances.

We used the {\em successor} representation, while the {\em set variable} or the {\em graph} representations should be
experimented extensively, possibly mixing different representations with tunneling constraints.

As future work we also plan to apply extensions of the proposed techniques in the Euclidean VRP and possibly to other similar problems. In a \ac{VRP} there is a fleet of vehicles that must reach all the
nodes of a graph; in each node the vehicle must load (or unload) goods. Each vehicle has a
capacity that must not be exceeded, so it is not possible to use a single vehicle (with only
one vehicle the \ac{VRP} boils down to the \ac{TSP}). It can be seen that in an optimal solution of
Euclidean \ac{VRP} there are no intersections in the path of each vehicle, even though there
may be intersections between the paths of different vehicles.

We are also planning to extend some of the techniques presented here in CLP to \ac{ASP}.

%\mg{Puoi mettere gli acknowledgement che erano in AAAI?}
\section*{Acknowledgments}
The author wish to thank Prof. Marco Gavanelli for supervising his PhD research activities and Joachim Schimpf for his help in the implementation of some of the constraints.
This work was partially supported by GNCS-INdAM.

\bibliographystyle{eptcs}
\bibliography{bibliography}
\end{document}